\documentclass{article}

\usepackage{microtype}
\usepackage{graphicx}
\usepackage{subfigure}
\usepackage{booktabs} 

\usepackage{hyperref}

\usepackage[accepted]{icml2023}

\usepackage{amsmath}
\usepackage{amssymb}
\usepackage{mathtools}
\usepackage{amsthm}
\usepackage{xspace}
\usepackage{url}

\usepackage{breakurl}
\usepackage[capitalize,noabbrev]{cleveref}

\theoremstyle{plain}

\theoremstyle{definition}

\theoremstyle{remark}

\newcommand{\sys}{GEAR\xspace}
\newcommand{\tinyskip}{\vspace{3pt}}
\newcommand{\mypar}[1]{\tinyskip\noindent\textbf{#1.}\xspace}

\usepackage{tikz}
\newcommand*\myc[1]{%
\scalebox{0.78}{\begin{tikzpicture}[baseline=-3pt]
  \node[draw,circle,inner sep=0.5pt, fill=black] {\textcolor{white}{\textsf{\textbf{#1}}}};
\end{tikzpicture}}}

\usepackage[textsize=tiny]{todonotes}

\icmltitlerunning{\sys: A GPU-Centric Experience Replay System for Large Reinforcement Learning Models}

\begin{document}

\twocolumn[
\icmltitle{\sys: A GPU-Centric Experience Replay System \\ for Large Reinforcement Learning Models}



\icmlsetsymbol{equal}{*}

\begin{icmlauthorlist}
\icmlauthor{Hanjing Wang}{equal,sjtu,dbl}
\icmlauthor{Man-Kit Sit}{equal,edin}
\icmlauthor{Congjie He}{edin}
\\
\icmlauthor{Ying Wen}{sjtu}
\icmlauthor{Weinan Zhang}{sjtu}
\icmlauthor{Jun Wang}{ucl}
\icmlauthor{Yaodong Yang}{pku}
\icmlauthor{Luo Mai}{edin}
\end{icmlauthorlist}

\icmlaffiliation{sjtu}{Shanghai Jiao Tong University}
\icmlaffiliation{dbl}{Digital Brain Lab}
\icmlaffiliation{edin}{University of Edinburgh}
\icmlaffiliation{ucl}{University College London}
\icmlaffiliation{pku}{Peking University}

\icmlcorrespondingauthor{Ying Wen}{ying.wen@sjtu.edu.cn}
\icmlcorrespondingauthor{Weinan Zhang}{wnzhang@sjtu.edu.cn}
\icmlcorrespondingauthor{Luo Mai}{luo.mai@ed.ac.uk}

\icmlkeywords{Machine Learning, ICML}

\vskip 0.3in
]



\printAffiliationsAndNotice{\icmlEqualContribution} 

\begin{abstract}

This paper introduces a distributed, GPU-centric experience replay system, \sys, designed to perform scalable reinforcement learning (RL) with large sequence models (such as transformers). With such models, existing systems such as Reverb face considerable bottlenecks in memory, computation, and communication. \sys, however, optimizes memory efficiency by enabling the memory resources on GPU servers (including host memory and device memory) to manage trajectory data. Furthermore, it facilitates decentralized GPU devices to expedite various trajectory selection strategies, circumventing computational bottlenecks. \sys is equipped with GPU kernels capable of collecting trajectories using zero-copy access to host memory, along with remote-directed-memory access over InfiniBand, improving communication efficiency. Cluster experiments have shown that \sys can achieve performance levels up to 6$\times$ greater than Reverb when training state-of-the-art large RL models. \sys is open-sourced at \url{https://github.com/bigrl-team/gear}.

\end{abstract}

\section{Introduction}

Recent breakthroughs in AI technologies have paved the way for training large sequence models~\cite{vaswani2017attention}, such as Gato~\cite{gato}, DB1~\cite{db1}, MAT~\cite{MAT} and GPT~\cite{brown2020language}, using a reinforcement learning (RL) approach~\cite{openai2023gpt4}. These models have set new benchmarks in complex decision-making such as game playing, robotics control, and question answering~\cite{gato, DT, MAT, trajectory-transformer}, recommender systems~\cite{geng2022recommendation, sima2022ekko} and AI-assisted generated content~\cite{ramesh2022hierarchical, shen2023flanmoe}. 

A pivotal component that makes RL viable for these large sequence models, or "large RL models," is the experience replay system. This system archives past experiences, organized as trajectories, for RL agents. These trajectories are gathered by RL policies through on-policy, off-policy, or offline methods~\cite{atari, mujoco}. For each training batch, the experience replay system chooses trajectories based on a certain strategy, such as first-in-first-out (FIFO), last-in-first-out (LIFO), weighted replay, or prioritized replay. These selected trajectories are then acquired by GPU-based training servers, executing multi-dimensional parallelism for sequence models~\cite{huang2019gpipe, deepspeed}.


Training large RL models using experience replay systems presents several challenges:
(i)~Such systems necessitate a substantial number of servers to store hundreds of terabytes of trajectories in memory~\cite{gato}. This requirement makes operating these experience replay systems exceedingly expensive.
(ii)~The complexity of trajectory selection strategies often escalates with the size of the batch~\cite{dota} and the total number of trajectories~\cite{gato}, leading to significant computational costs.
(iii)~Training large RL models invariably involves using large batch sizes, which can result in collecting massive trajectories over the network per training iteration. This process imposes exorbitant communication costs.
Given these challenges, it is crucial to consider the storage, computational, and communication costs incurred by an experience reply system when supporting large RL models with experience replay systems.

Existing experience replay systems, unfortunately, fall short in fully addressing the aforementioned challenges. Most of these systems, such as RLlib~\cite{rllib}, RLZoo~\cite{ding2021efficient}, stable-baselines~\cite{stable-baselines}, rlpyt~\cite{rlpyt}, tianshou~\cite{tianshou}, TorchOpt-RL~\cite{liu2022theoretical, ren2022torchopt}, sample factory~\cite{sf}, and envpool~\cite{envpool}, are incorporated as part of single-server RL frameworks and fail to offer distributed trajectory storage, selection, and collection.
The recent development in distributed experience replay systems, exemplified by Reverb~\cite{reverb}, allows for storing trajectories on memory-optimized servers. Reverb utilizes CPU processors for trajectory selection and collection during training. However, it remains significantly inefficient: to store massive trajectories, Reverb demands the deployment of numerous additional servers, thereby incurring high storage costs. It also struggles with performance issues when executing selection tasks on a large number of trajectories due to the restricted parallelism of CPU processors.
Furthermore, Reverb employs RPC libraries (specifically, gRPC) to collect trajectories, leading to excessive memory copies and data serialization. This methodology exhibits low communication efficiency.

In this paper, our objective is to design an experience replay system that can be effectively employed in training large RL models. We have observed that the training servers typically possess vast memory, computation, and communication resources. These servers boast a large amount of host memory (usually ranging from 1 to 4 terabytes) and multiple GPUs (up to 8). These GPUs have high-bandwidth memory and are interconnected with high-bandwidth connectivity including NVlink and InfiniBand.
Our primary design strategy, therefore, involves leveraging these training servers to:
(i)~Store and manage trajectories in their host memory, eliminating the need to use additional servers for trajectory storage.
(ii)~Speed up trajectory selections with the help of distributed GPUs.
(iii)~Ensure efficient collection of trajectories using InfiniBand, which provides high bandwidth and zero-copy direct access to remote data.

To realize the above idea, we design and implement \sys, a novel distributed GPU-centric experience replay system. The design of \sys makes the following contributions in scaling the training of large RL models:

\mypar{(1) Trajectory management on training servers} \sys can divide trajectories into shards and allocate these shards across distributed training servers. The trajectory sharding strategy enhances data locality by factoring in the topology of pipeline parallelism and trajectory priorities during selection. Moreover, \sys incorporates an optimized trajectory storage format wherein trajectory fields selected together are placed in continuous memory, thereby maximizing data locality and bandwidth utilization.

\mypar{(2) GPU-optimized distributed trajectory selection} 
\sys allows various trajectory selection strategies to benefit from distributed GPUs. It realizes centralized trajectory selection which guarantees deterministic selection results when using distributed GPUs. It further realizes decentralized trajectory selection, thus parallel GPUs can contribute partial trajectory selection results, significantly improving the efficiency of selecting trajectories in extremely large datasets (e.g., those with 100s TB trajectories).

\mypar{(3) GPU-centric trajectory collection}
\sys has optimized GPU kernels that maximize the communication efficiency in collecting trajectories in GPUs. For trajectories in local host memory, the GPU kernels use zero-copy 
directed-memory-access (DMA), bypassing CPUs and avoiding data copies and serialization. For trajectories on remote servers, the GPU kernels can launch RDMA send/receive to retrieve the trajectories over InfiniBand. 

We evaluated \sys in a 24-GPU cluster with state-of-the-art large RL models including Gato~\cite{gato} and MAT~\cite{MAT}. Experimental results show that \sys achieves up to better 6x performance (35 GB/s throughput in collecting trajectories) compared to the state-of-the-art Reverb (6 GB/s) with a wide range of configurations (different trajectory sizes, different models and different datasets).


\section{Background and Motivation}

In this section, we describe the background and motivation for designing \sys.


\subsection{Reinforcement learning for large sequence models}

Recent studies have demonstrated the considerable benefits of incorporating RL into large sequence model training. Typically, an RL-based sequence model training system comprises (1) actors, which generate trajectories online through simulation environments such as Atari~\cite{dqn}, Mujoco~\cite{mujoco}, and Google Football~\cite{gf}, and (2) learners, which persistently select a batch of trajectories to train a deep neural network, herein referred to as the model~\cite{rllib}. 

Furthermore, trajectories can also be produced by empirical expert policies offline, allowing large sequence models to mimic or even surpass the performance of these policies. Real-world datasets are another valuable source of trajectories~\cite{trajectory-transformer}, such as D4RL~\cite{d4rl}, Minecraft~\cite{minedojo}, and robot manipulation~\cite{bcz}. 

Current large sequence models possess billions or even trillions of parameters. To counter the memory restrictions of a single GPU, developers must employ multi-dimensional parallelism. This involves partitioning and replicating the model, with different model partitions being executed using multi-dimensional parallelism~\cite{zheng2022alpa}: a combination of data parallel~\cite{mai2020kungfu}, model parallel~\cite{rajbhandari2020zero}, and pipeline parallel approaches~\cite{huang2019gpipe}.

A pivotal component in large RL model training systems is the experience replay system, which manages massive online and offline produced trajectories. This system enables the model training system to select a batch of trajectories continuously, based on a specific trajectory selection strategy. Following trajectory selection, the training system computes gradients to refine the models. For online RL, the models also return actions to the simulation environments.

\subsection{Challenges for experience replay systems}

Training large RL models, while promising, presents several significant challenges today. The key challenges include:

\mypar{(1) High storage costs incurred by trajectory storage}
The need for extensive datasets to train large RL models has intensified, incorporating massive offline-generated trajectories by various empirical models and trajectories collected over extended periods from parallel environment simulators. For instance, the dataset used to train the DB1 model~\cite{db1} encompasses 110 TBs (over 350 billion tokens), mirroring the datasets employed for training the DeepMind Gato~\cite{gato}. These trajectories must be retained in server memory for subsequent selection and transport to training servers. However, since commodity servers only offer up to a few TBs of memory, storing the entire dataset necessitates hundreds of servers, leading to substantial memory costs. For example, a memory-optimized server with 1TB of memory in a public cloud can cost several dollars per hour. Thus, using multiple servers to store a dataset of 110 TBs could lead to storage costs amounting to several thousand dollars per hour.

\mypar{(2) High computation costs associated with trajectory selection}
The process of selecting trajectories for the training of large RL models presents considerable computational challenges. Even with the support of auxiliary data structures~\cite{reverb}, such as max heaps and prefix-sum trees, to facilitate the selection process, the need for large batch sizes and the high volume of trajectories often required in training results in a significant number of iterations necessary to complete the selection. For instance, using a prefix-sum tree to conduct priority-based trajectory selection has a time complexity of $O(B \times log N)$, where $N$ is the total number of trajectories and $B$ is the batch size. When $B$ is 1 million and $N$ is 1000 million, the prefix-sum tree still requires more than 10 million iterations to complete trajectory selection. This computational demand can be burdensome for CPUs.


\mypar{(3) High communication costs due to trajectory collection}
Training large RL models often necessitates the collection of an extensive amount of sizeable trajectories within a training batch. A trajectory can vary in size from hundreds of kilobytes to megabytes. This is attributable to (i) the use of large RL models in long-term planning tasks, such as the game of Go~\cite{alphago}, which often involve trajectories tied to thousands of timestamps~\cite{alphastar,dota}, and (ii) the fact that large RL models may also employ trajectories with multi-modal data~\cite{nair2022learning}, including text, images, and videos~\cite{minedojo}. The collection of such a vast number of large trajectories places a significant demand on communication bandwidth at the training servers. For instance, gathering one million trajectories, each sized at 100 KBs, necessitates the transfer of more than 100 GBs of data. This rate poses a challenge for current communication technologies~\cite{mai2015optimizing}, such as Ethernet and the PCIe bus, which respectively offer 10-40 Gbps and 32 GBs of bandwidth~\cite{koliousis12crossbow}.




\subsection{Limitations of existing systems}

Existing experience replay systems such as RLlib~\cite{rllib}, stable-baseline~\cite{stable-baselines}, rlpyt~\cite{rlpyt}, tianshou~\cite{tianshou}, sample factory~\cite{sf}, and envpool~\cite{envpool} have been predominantly designed for relatively smaller RL models and their implementations are confined to single-server contexts. Consequently, they are not equipped to handle the distributed trajectory storage, selection, and collection necessary for training large RL models.

Recently, DeepMind presented Reverb~\cite{reverb}, a distributed experience replay system and a key component of their Acme research framework~\cite{acme}. Reverb employs a collection of memory-optimized servers for trajectory storage and relies on CPU processors for trajectory selection. Subsequently, the chosen trajectories are conveyed to training servers via gRPC, which uses network sockets.

However, despite its strengths, Reverb falls short of fully tackling the challenges inherent in training large RL models. To store datasets featuring hundreds of terabytes of trajectories, Reverb necessitates a considerable number of CPU servers, leading to elevated storage costs. Furthermore, the limited parallelism offered by CPU processors translates to lengthy selection times (spanning several seconds), which is starkly contrasting to the time required to complete a batch of GPU training (typically in the order of hundreds of milliseconds)~\cite{megatron}. In addition, the use of network sockets for trajectory transfer involves multiple data copies, such as moving trajectories from user space to the operating system kernel, and necessitates the serialization and deserialization of trajectory data. This restricts Reverb's throughput to a few gigabytes per second, which is an order of magnitude less than the throughput (hundreds of gigabytes per second) required by large RL models.

\section{\sys Design and Implementation}

In this section, we delve into the design and implementation of \sys. Our design approach stems from an observation that modern training servers for large RL models are typically equipped with substantial memory resources (such as terabyte-scale server memory and SSDs), computation accelerators (for instance, 8-16 GPU devices), and high-bandwidth networks (including 40 GB/s InfiniBand and 600 GB/s GPU NVLink).

We utilize these training servers to: (i)~leverage the host memory for storing and managing trajectories, thereby eliminating the need for additional storage servers, (ii) employ parallel CUDA kernels to hasten the process of trajectory selection, and (iii) incorporate GPU-centric data communication methods to gather trajectories through PCIe and high-bandwidth networks.


\subsection{Overview}

\begin{figure}[t!]
  \centering
  \includegraphics[width=\linewidth]{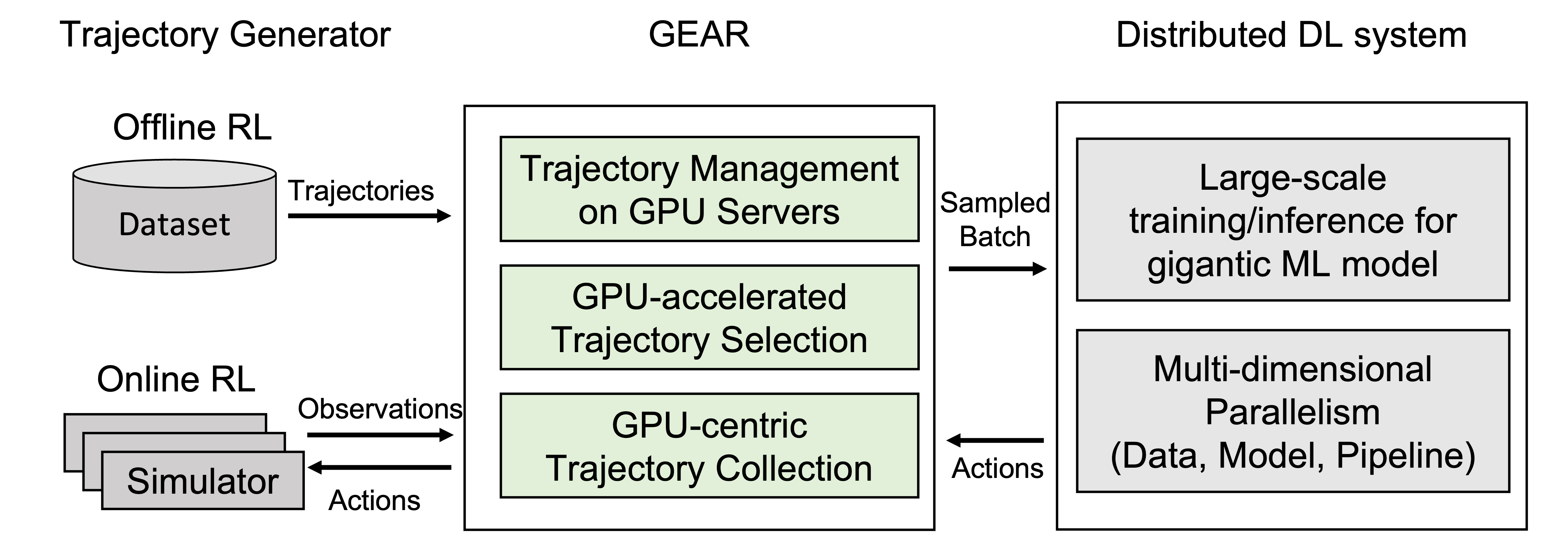}
  \caption{Overview of \sys}\label{fig:overview} 
\end{figure}

Figure~\ref{fig:overview} offers an overview of \sys. Within \sys, trajectory generators both write trajectories to and read trajectories from the system. \sys is responsible for managing and transporting these trajectories between the trajectory generator and the distributed DL system. Ultimately, \sys operates in tandem with a distributed DL system that trains and infers large sequence models using multi-dimensional parallelism.

\sys is designed to support two training scenarios for large sequence models: \emph{offline RL} and \emph{online RL}. (i)~In the offline RL scenario, \sys ingests trajectories from training datasets that have been pre-collected offline. These trajectories are then inserted into \sys's trajectory storage. The DL system subsequently samples trajectory batches from \sys in order to update the model parameters. (ii)~In the online RL scenario, \sys employs environment simulators to generate observations. These observations are recorded in \sys and subsequently transferred to the DL system, which responds with actions. These actions are relayed back to the simulators to continue the simulation process.






\subsection{Trajectory management on training servers}

Enabling trajectory management on distributed training servers exhibits several unique challenges: (i)~The training servers realize pipeline parallelism, and based on their assigned roles in the pipeline, the servers need to be assigned with different shards of trajectories (e.g., only the first server in a training pipeline will need to collect trajectories), (ii)~The trajectory sharding scheme must optimize data locality (that is, the majority of the trajectories collected by the servers reside in local memory), and (iii)~The trajectory storage format needs to optimize for the unique data access pattern of trajectories (that is, trajectories are collected based on their priorities, a range of timestamp and a sub-group of fields specified by the developers of large RL models).

\begin{figure}[t!]
  \centering
  \includegraphics[width=\linewidth]{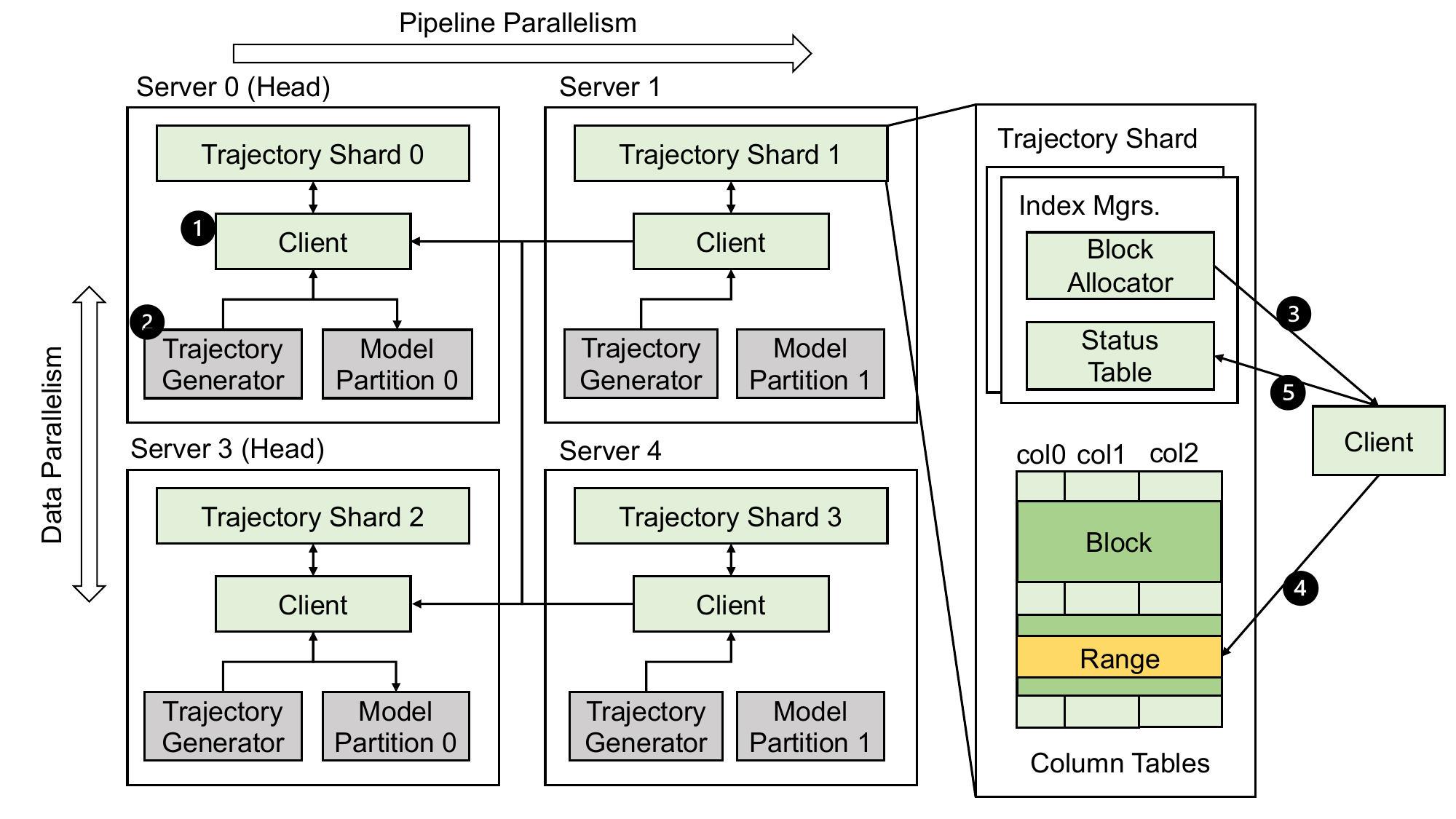}
  \caption{Trajectory management on distributed training servers}\label{fig:store-design} 
\end{figure}

\mypar{Pipeline-aware trajectory sharding} Figure~\ref{fig:store-design} gives an overview of the management system design in pipeline parallel training servers. In this figure, we have 4 training servers, where server 0 and 3 are the head servers in pipeline parallelism. The trajectory storage is partitioned into equal-sized shards to allow for storing very large amounts of trajectories. Each shard is hosted by one machine. As each GPU server contains up to several TB of system memory, the shards are stored in the system memory of the GPU servers. 

The client (\myc{1}) can write to the local shard and read from the remote shards. Trajectory generators (\myc{2}) write trajectories to the store through the client, which can be either offline (from datasets) or online (from simulators). The placement controller decides the placement of the shards to improve data locality. The placement controller considers both the type of trajectory generator and parallelism. With pipeline parallelism, only the head servers which run the input layer needs to read the shards. Therefore, for offline generators, we place the data with a higher probability (i.e. higher priority) to be selected in the head machines. For an online generator, the trajectories are written to the local shard where the trajectories are generated to reduce write overheads.

\mypar{Shard storage format} 
Clients typically need to read only certain fields of trajectories (such as observation, action, or reward) and write to the storage field by field. To facilitate this access pattern, we chose a column-based storage system, which stores the same type of fields together in a consecutive manner. Each shard is composed of two main components: column tables and index managers. These components are allocated in shared memory, allowing direct access by client processes without the overhead of inter-process communication (IPC).

A column table holds a specific field type of a trajectory. Since the shape of the fields is defined by the users at creation, the tables are implemented as continuous arrays divided into blocks of equal size. Each block represents a field of a trajectory and serves as the basic unit of read/write operations. The fields within a block are stored as a sequential chain of flattened tensors. The storage capacity is defined by the number of blocks in a table, and all tables share the same capacity, representing the maximum number of trajectories that a shard can accommodate.

\mypar{Index manager} The Index Manager comprises a Block Allocator and a Status Table. Each block within a table can be uniquely identified by an index, with a row of blocks (i.e., blocks sharing the same index across different tables) forming a complete trajectory.

The Block Allocator is tasked with block allocation and the subsequent status updates. To manage the indices of free blocks – those that are either empty or contain outdated trajectories awaiting eviction – the allocator employs a single queue. The processes of allocation and release involve dequeue and enqueue operations, respectively. The Status Table, on the other hand, maintains essential statistics such as priorities and timestamps that reflect the usability of indices and their associated trajectories. The priorities of indices that represent ongoing or evicted trajectories are set to zero to avoid their selection. When allocation or eviction events occur, the updates on block status are consolidated and committed to the Status Table.

\sys utilizes a \emph{multi-controller} architecture, where client programs are duplicated and run to manage exclusive portions of hardware resources, such as GPU devices. For preventing data corruption, \sys directs clients and index managers to periodically synchronize with each other, ensuring clients have up-to-date information on block allocation.

In line with the principles of a multi-controller architecture, trajectory shards are further divided and managed by individual local index managers. Each local index manager is tied to a specific client process to minimize local synchronization overhead. This arrangement enables the Index Manager to smoothly integrate with existing deep learning libraries, including Torch-DDP and DeepSpeed.

\mypar{Trajectory insertion and deletion} To insert a trajectory into the store, the client initiates an \texttt{allocate} operation, which generates a buffer containing memory views of the blocks. As demonstrated in Figure~\ref{fig:store-design}, this operation involves several steps: \myc{3} The client requests a free index from the block allocator. \myc{4} The client retrieves the memory locations of the indexed blocks from each column table. These memory locations are encapsulated into a buffer, which the client then fills with trajectory data. After writing the blocks, the client \myc{5} \texttt{commit}s the buffer, triggering an update in the Status Table. This update designates the index as available for selection. All write operations are performed in place to avoid unnecessary memory copying.

Releasing a row requires the block allocator to enqueue the freed index and designate it as unavailable for selection. This occurs when either (1) the block allocator receives a request but no index is available, or (2) the number of selectable indices reaches the preset maximum capacity. In these scenarios, the allocator automatically selects a victim index to be released according to a user-defined removal strategy (e.g., FIFO, LIFO).

\subsection{Trajectory selection with distributed GPUs}

\sys facilitates efficient trajectory selection using distributed GPUs on training servers. It addresses key challenges, such as (i) ensuring deterministic selection results for consistent training outcomes of large RL models; (ii) maintaining consensus among all participants on the selected indices to avoid data corruption; and (iii) controlling communication overheads for effective scaling.

\begin{figure}[t!]
  \begin{minipage}[t]{.23\textwidth}
    \centering
    \includegraphics[width=1.5in]{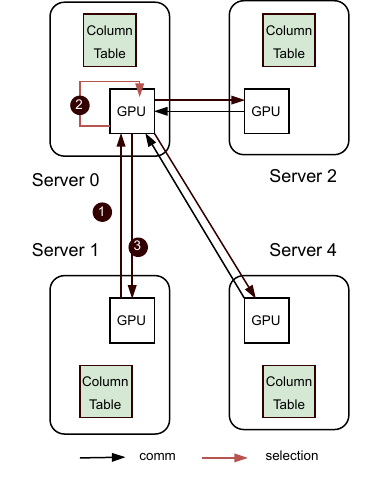}
    \centerline{(a) Centralized selection}
  \end{minipage}
  \begin{minipage}[t]{.23\textwidth}
    \centering
    \includegraphics[width=1.5in]{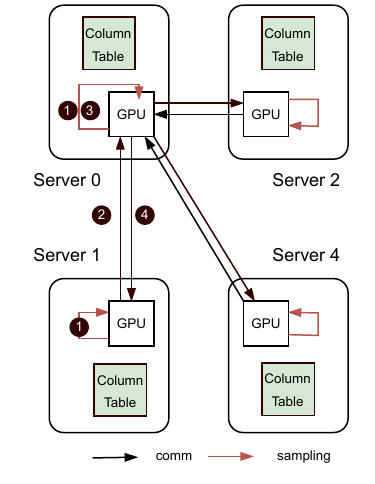}
    \centerline{(b) Decentralized selection} 
  \end{minipage}
  \label{fig:trajectory_selection}
  \caption{Centralized and decentralized trajectory selection }
\end{figure}

\mypar{Centralized trajectory selection}
Figure 2(a) illustrates the process of centralized trajectory selection, which consists of the following steps:
\myc{1} The GPU in the central server conducts a global gathering operation to collect selectable indices and weights from all servers.
\myc{2} A selection algorithm then picks from these collected indices and generates a list of selected indices.
\myc{3} The central server broadcasts this list of selected indices to the other servers.

\sys supports both uniform sampling and weighted sampling as centralized selection algorithms. We've implemented an optimized CUDA kernel for weighted sampling. This kernel first computes a prefix sum array using the decoupled look-back algorithm and then performs binary searching to locate the bins associated with uniformly generated random numbers. The prefix sum calculation and the searching procedure require $ \frac{k \times \log{N}}{s}$ steps, where $s$ is the degree of parallelism, $k$ is the number of samples, and $N$ is the sample size.




\mypar{Decentralized trajectory selection} \sys{} provides FIFO and TopK selection implementation, which are deterministic in decentralized selection scenarios. Therefore, all servers can perform a local scan to generate $k$ samples before the global gathering operation, which can significantly reduce the communication overhead from $O(n)$ to $O(mk)$ when $k\ll n$, where $m$ is the parallel world size of servers.

As depicted by Figure 2(b), the decentralized trajectory selection in \sys will first run GPU kernels in each GPU to compute the partially selected trajectories (e.g., the top-K priority trajectories) (\myc{1}). The partially selected trajectories will be sent to the central GPU (\myc{2}) to compute the global top-K selected trajectories. The global top-K selected trajectories will be broadcast to all GPUs and allow these GPUs to retrieve their trajectories (\myc{3}). 

\subsection{GPU-centric trajectory collection}

\begin{figure}[t]
  \centering
  \includegraphics[width=\linewidth]{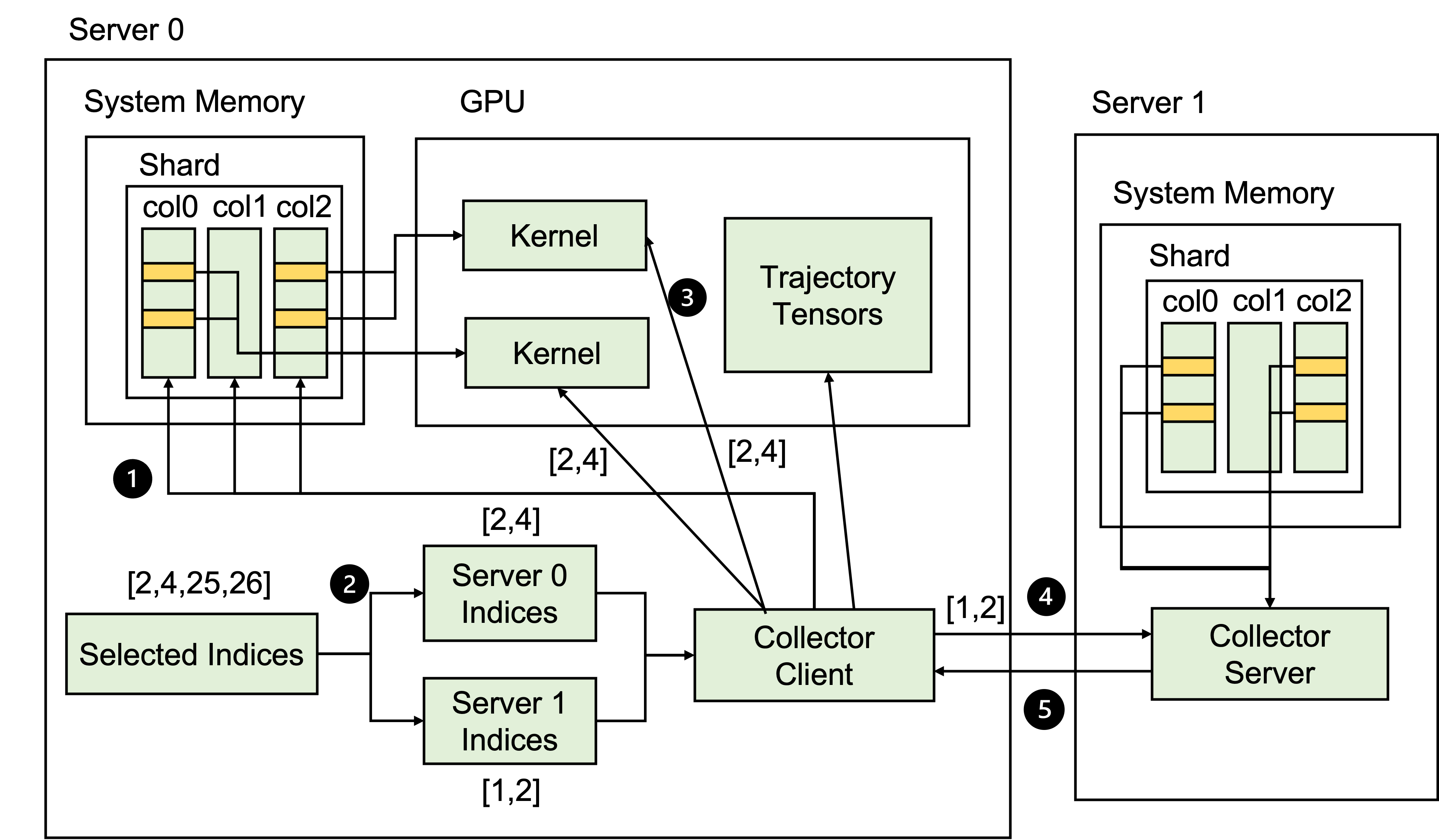}
  \caption{Trajectory collection on distributed GPUs}\label{fig:collection} 
\end{figure}

\sys aims to eliminate data copies and serialization, utilizing high-bandwidth networks for GPUs, thereby reducing latency when collecting trajectories for training servers. However, we identify two key research gaps: (i) Modern GPUs can directly read from shared memory, negating the need for data copying and serialization over CPU-managed memory. This capability is yet to be fully exploited in current machine learning frameworks. (ii) GPU servers often have InfiniBand, providing high bandwidth and efficient hardware-assisted data copy and serialization. However, in existing machine learning frameworks: InfiniBand is primarily used for synchronizing the gradients for RL models, leaving it underutilized in collecting trajectories.


\mypar{Index translation}
As the selected list of indices is global indices, it is translated into lists of local indices for each server. Since the shards have equal size, it is simple to determine if an index belongs to which machine by dividing the index by the capacity. The translated lists of local indices are sent to the collector.

\mypar{Trajectory collection}
GEAR enables GPUs to directly collect trajectory in host memory by facilitating one-sided data accesses. The collector consists of a collector client and a collector server. For local collection, the collector utilizes the zero-copy access feature of NVIDIA GPUs, which allows GPU threads directly access the host memory without the help of the CPU. Compared to the more common DMA-based data transfer, it is more suitable for sparse data accesses~\cite{tan2023quiver}, as it allows GPUs to send more fine-grained memory requests to the system memory directly through PCIe. To enable zero-copy, the memory of the tables is pinned to page-lock the table data (\myc{1} in Figure~\ref{fig:collection}). The collector client launches one CUDA kernel per table to collect trajectories from host memory to GPU memory. For remote trajectory collection, the collector utilizes NCCL to copy trajectories from remote shards through InfiniBand.

\mypar{Collection example}
For example, in Figure~\ref{fig:collection}, a client wants to collect field \texttt{"col0"} and \texttt{"col1"} of indices 2, 4, 25, 26 from the store, it calls \texttt{collect([2, 4, 25, 26], ["col0", "col1"])}. The indices are translated into [2, 4] and [1, 2] which are the local indices of server 0 and server 1 respectively (\myc{2}). The collector client launches 2 CUDA kernels to read locally from each of the requested tables (\myc{3}). The client sends [1, 2] to the collector server in server 1 to request the trajectories remotely (\myc{4}). After the collector server collected the trajectories, it sends them to the collector client (\myc{5}). The local and remote collected trajectories are concatenated and converted to Pytorch Tensor.

\subsection{Implementation details}

The current implementation of \sys comprises 4,300 lines of C/C++ code and 2,000 lines of Python code. \sys supports large RL models authored in PyTorch and parallelized by DeepSpeed, a popular distributed training and inference library for sizable transformers. It facilitates the import of offline PyTorch and TensorFlow datasets and accommodates data generated by a diverse assortment of environment simulators, such as Atari, Google Football, Starcraft II, and Robot Arms.

\mypar{Multi-framework support} \sys is designed as a framework-independent library, enabling its integration with various machine learning frameworks. Currently, it exposes trajectory data via the PyTorch Tensor interface, facilitating seamless interaction and data exchange between \sys and PyTorch models. 

Moreover, \sys is incorporated into DeepSpeed to support multi-dimensional parallelism when training large sequence models. Distributed communication and additional data placement optimizations that hinge on pipeline parallelism can be facilitated by DeepSpeed's parallel topology unit.  

\mypar{Failure recovery}
\sys allows for trajectory shards to be checkpointed on local SSDs. Developers frequently trigger checkpointing processes at data epoch boundaries which are aligned with model parameter checkpoint boundaries~\cite{mai2020kungfu}. Since \sys is incorporated into the DeepSpeed framework, it relies on DeepSpeed to detect failures and recover trajectory shards.

\section{Experiments}



In this section, we outline our experiment settings and results to assess the performance and accuracy of \sys.

\mypar{Cluster} We conduct benchmarking tests and RL training tasks on a three-server cluster, each being a standard NVIDIA DGX-A100 server. Each server houses dual AMD Rome 7742 processors, eight NVIDIA A100 GPUs, 1000GB of RAM, and high-bandwidth NVLink and IB connections. This setup enables us to evaluate \sys's scalability for distributed training and its effectiveness as a seamless substitute for the existing RL codebase without impacting model convergence.

\mypar{Baseline} We choose Reverb, a widely adopted distributed experience replay system, as our comparison baseline. We conduct extended benchmarks to measure the average sampling throughput and compare \sys's performance against Reverb. It's worth noting that Reverb relies heavily on TensorFlow, complicating its direct integration with our PyTorch-based benchmark kits. To tackle this, we develop mock clients for both \sys and Reverb that merely collect and instantly discard data, bypassing the conversion procedure and neutralizing potential benchmarking biases due to different data interfaces.



\subsection{Trajectory throughput}

In our study, we compare the trajectory selection throughput between \sys and Reverb. This involves the implementation of clients that concurrently generate selection requests and gather data from a central server process—a common practice in Single-Program-Multiple-Data (SPMD) parallel programs.

In the Reverb setup, the server process is hosted on server 0. In contrast, for \sys, global aggregations and selection are managed by GPU 0 on server 0. Additionally, each GPU within \sys is allocated to a unique client process. All these processes then form a data parallel group, enabling synchronized data retrieval and consumption.
The purpose of this experimental setup is to push both \sys and Reverb to their maximum potential throughput. This offers a rigorous assessment of their performance when tasked with handling heavy workloads.

\begin{figure}[t!]
    \centering
    \includegraphics[width=1.0\linewidth]{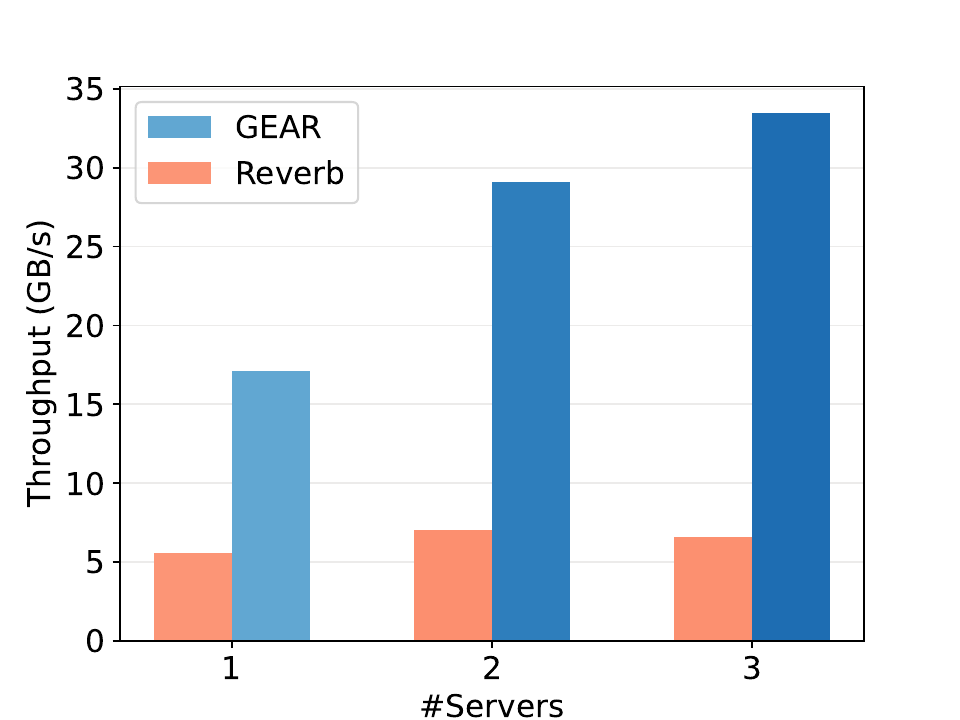}
    \caption{End-to-end throughput comparison with Reverb}
    \label{fig:throughput}
\end{figure}

The results depicted in Figure~\ref{fig:throughput} illustrate that, under single-node settings, Reverb reaches its peak single-node throughput at 6.73~GB/s with 64 parallel clients. Conversely, \sys{} achieves a maximum throughput of 17.1~GB/s with only 8 clients involved in globally synchronized sampling loops.

In multi-server experiments, \sys{} attains a total throughput of 29.1~GB/s with 16 clients spread across two nodes, and a total throughput of 33.0~GB/s with 24 clients spanning three nodes. This demonstrates that \sys exceeds Reverb's performance by a factor of five.

We observed that \sys's total throughput increases more slowly than the scale, particularly when expanding from two nodes. This is primarily due to the strict global synchronization policy that was adopted to align with Reverb's behavior. This effect could be mitigated by implementing parallel sampling techniques such as reservoir sampling and divide-and-conquer sampling. These methods can be executed prior to global synchronization, reducing communication overhead. However, it's important to note that these techniques could have side effects in distributed scenarios. For instance, they might provide less strict theoretical guarantees or might adversely affect the reproducibility performance of downstream algorithms. Therefore, we leave the decision to the user as to whether or not to implement these techniques.

\subsection{Batch size}




\begin{figure}[t!]
    \centering
    \includegraphics[width=1\linewidth]{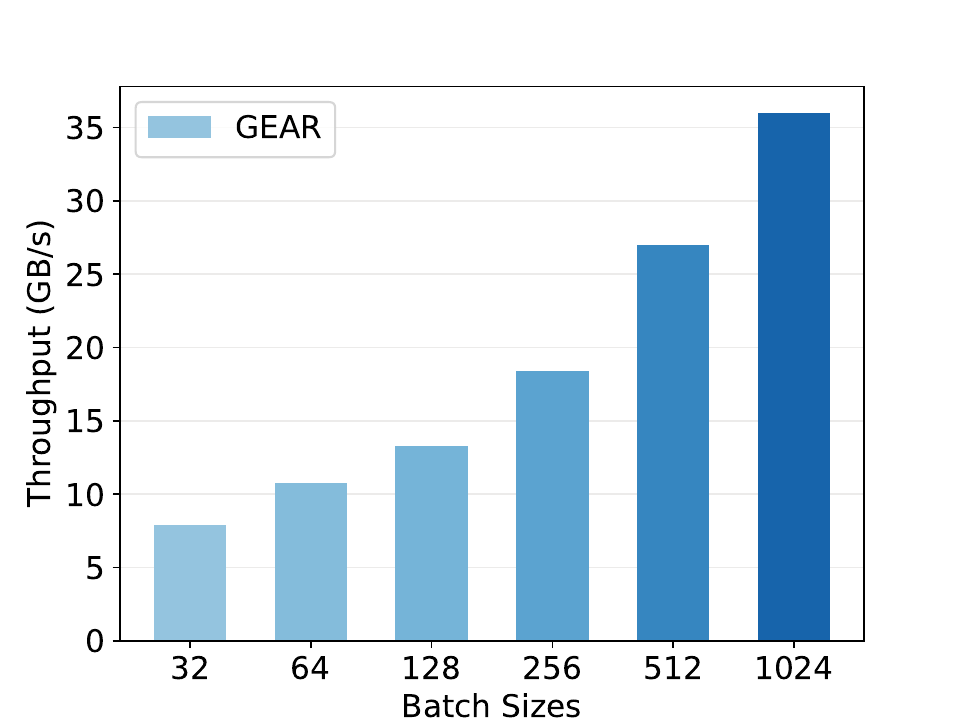}
    \caption{Trajectory collection throughput with varied batch sizes}
    \label{fig:diff_batchsize}
\end{figure}

In our experiment, we investigate the impact of varying batch size~\cite{mai2019taming} on both the computation cost and the communication cost of trajectory selection and collection in \sys. We carried out a set of tests with GEAR using a range of batch sizes: 32, 64, 128, 256, 512, and 1024. The results of this experiment are presented in Figure~\ref{fig:diff_batchsize}. As observed from the figure, \sys can achieve a linear scalability in trajectory collection performance with increasing batch sizes: its throughput starts at 8 GB/s using a small batch size 32 and the throughput jumps to 36 GB/s when a large batch size of 1024 is employed. This showcases GEAR's ability to effectively scale its performance with larger batch sizes.


\subsection{Model convergence}

We evaluate the correctness of trajectory selection of \sys by showing the convergence results of using \sys for training two popular large RL models: Gato~\cite{gato} and MAT~\cite{MAT}. The former covers the scenarios where trajectories are produced offline and the latter covers the cases where trajectories are produced by environment simulators online.

\mypar{Gato} We implement the GATO model using PyTorch and reproduce similar loss performance consistent with what they reported in their paper. The GATO model has 1 billion parameters and it is trained with a dataset that has 100 TBs trajectories. We report a minimal GATO experiment and convergence performance with \sys in the D4RL mujuco hopper task pretrained with the D4RL expert dataset. As we can see from Figure~5, the episode length convergence to $1k$ and episode return converged to $3000$
These results show that \sys has been correctly integrated with existing pipeline parallelism libraries and it can correctly select trajectories that make SOTA RL models to converge. 

\begin{figure*}
  \begin{minipage}[t]{.5\textwidth}
    \centering
    \includegraphics[width=3in]{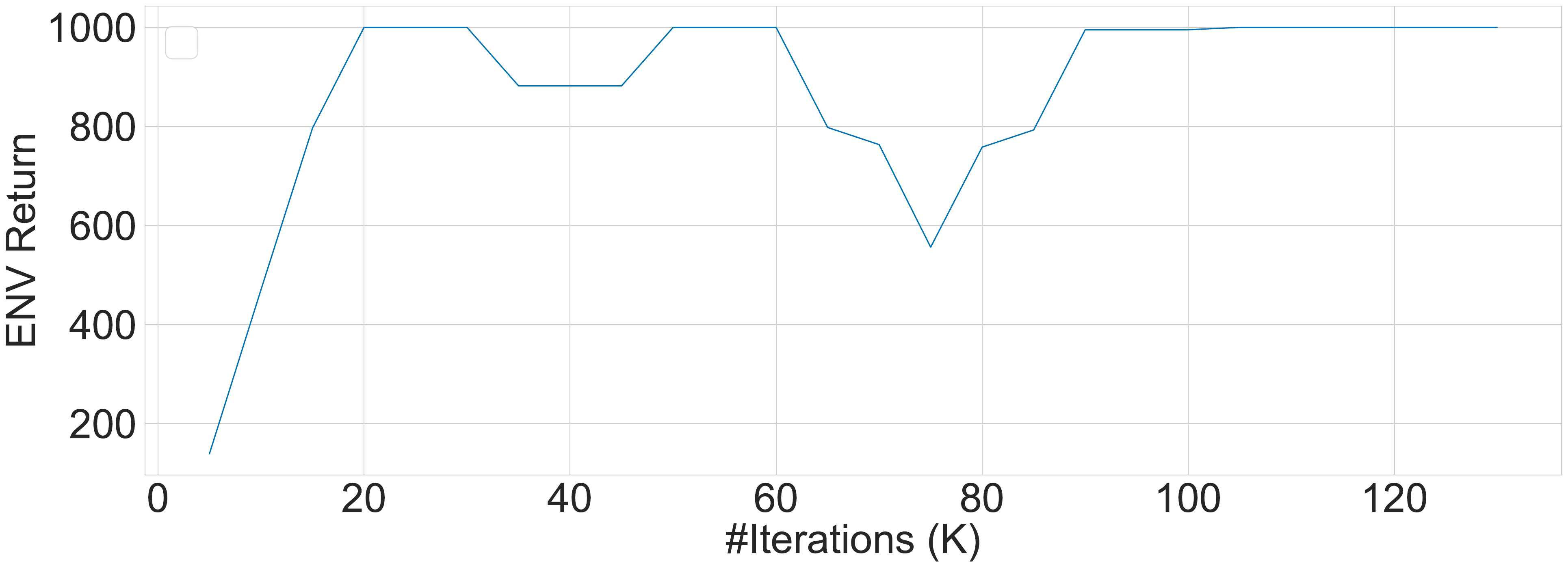}
    \centerline{(a) Episode length}
  \end{minipage}
  \begin{minipage}[t]{.5\textwidth}
    \centering
    \includegraphics[width=3in]{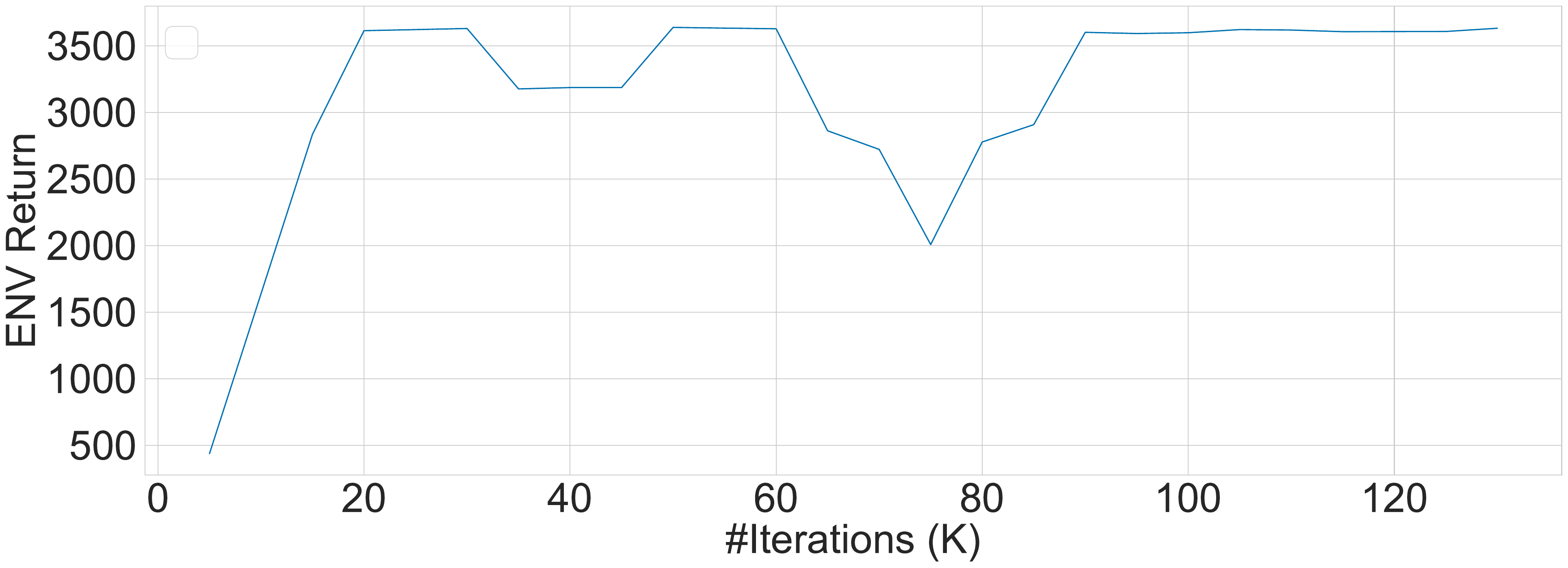}
    \centerline{(b) Episode return} 
  \end{minipage}
  \label{fig:gato}
  \caption{Convergence experiment of GATO with \sys. The x-axis is the number of iterations and the y-axis is the environment return.} 
\end{figure*}

\begin{figure*}[t!]
  \begin{minipage}[t]{.33\textwidth}
    \centering
    \includegraphics[width=2in]{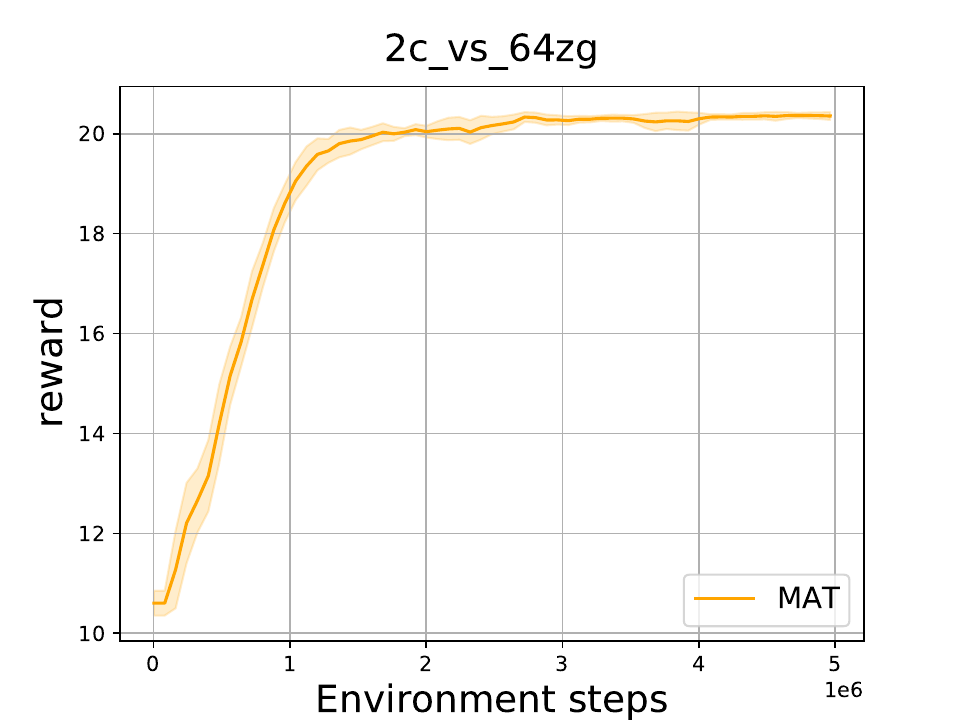}
    \centerline{(a) $2c_vs_64zg$}
  \end{minipage}
  \begin{minipage}[t]{.33\textwidth}
    \centering
    \includegraphics[width=2in]{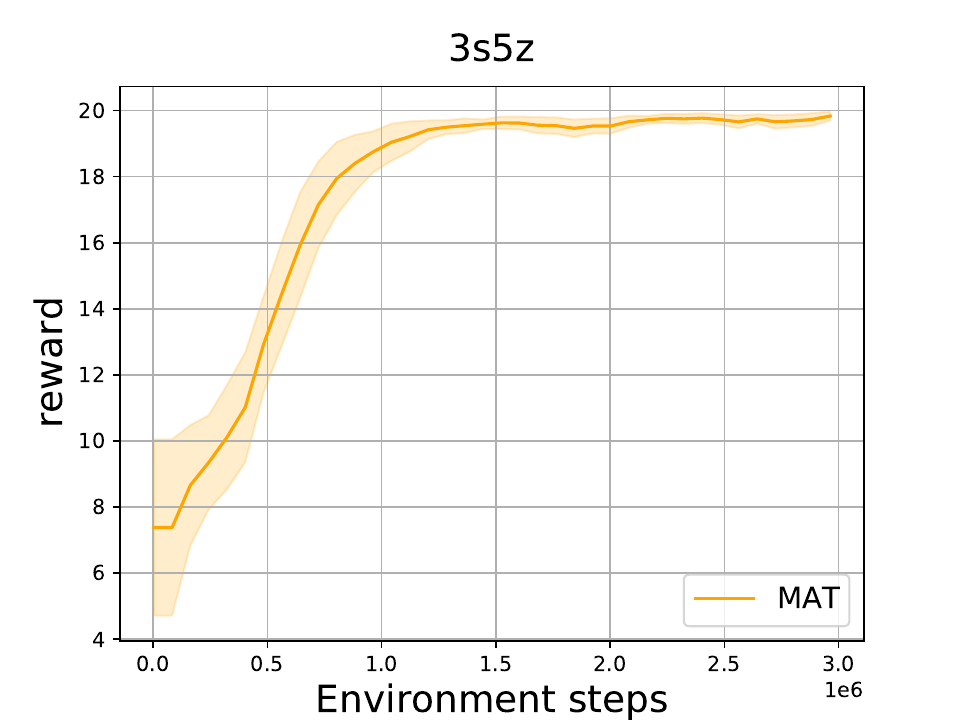}
    \centerline{(b) $3s5z$} 
  \end{minipage}
  \begin{minipage}[t]{.33\textwidth}
    \centering
    \includegraphics[width=2in]{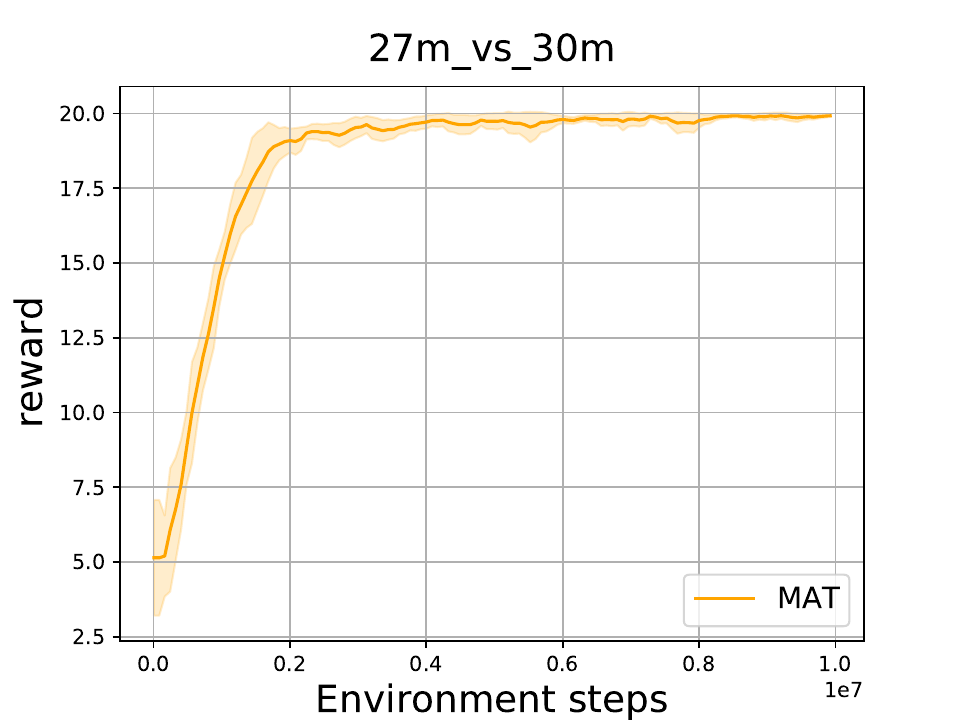}
    \centerline{(c) $27m_vs_30m$} 
  \end{minipage}
  \label{fig:MAT}
  \caption{Convergence experiment of MAT with \sys. The x-axis is the environment steps and the y-axis is the reward.}
\end{figure*}

\mypar{MAT} We also implement the MAT model, the state-of-the-art online multi-agent model that uses transformers as the backbone. The muli-agent RL model actually has a more challenging requirement for trajectory selection. Through this experiment, we show that \sys enables many multi-agent researchers to work on large RL models. The MAT model can be scalable to larger sizes flexibly. In this case, to illustrate \sys does not affect model convergence, we use a MAT model of $1$ block of hidden size $64$, which is aligned  We report the MAT performance with \sys in the StartCraftII tasks, namely $2c\_vs\_64zg$, $3s5z$, $27m\_vs\_30m$, which are correspondingly the "easy", "hard" and "super hard" tasks in SC2 environments. As we can see from Figure~6, \sys integrates MAT converged to $20$ in terms of environment returns, which is the desired convergence result of SC2.

\subsection{Performance Breakdown}

To provide a holistic understanding of \sys's performance, we carry out evaluations to analyze the individual contributions of each component and to quantify the enhancements they offer:

\mypar{Computation} We contrast GPU trajectory sampling with its CPU equivalents used in Reverb. Our results indicate that distributed GPU sampling is typically 100-1000 times quicker than the CPU versions.

\mypar{Memory access} We assess the GPU kernel that enables direct access to trajectories in host memory. This exhibits an impressive throughput of 17 GB/s, significantly outperforming the 2 GB/s achieved when a CPU thread is used to orchestrate data movement, as is the current practice in PyTorch and TensorFlow. This amounts to a performance increase by a factor of 8.5.

\mypar{Communication} With respect to InfiniBand-based distributed trajectory collection, we juxtapose \sys and Reverb, which employs gRPC for trajectory communication. The evaluation reveals that GEAR exhibits superior scalability when compared to Reverb. When utilizing four machines, GEAR reaches a communication throughput of 35 GB/s, outpacing Reverb, which caps at 5 GB/s.

\section{Conclusion}

This paper presents \sys{}, a distributed GPU-centric experience replay system designed to enhance the efficiency of large RL model training. \sys explores a novel design approach, employing a GPU-centric architecture for experience replay systems. It stores trajectories on distributed training servers, utilizes distributed GPUs to expedite trajectory selection, and enables GPUs to efficiently gather trajectories through directed memory access technologies. Our experimental evaluations demonstrate that \sys{} significantly enhances experience replay performance compared to leading systems: Reverb. We anticipate \sys playing a pivotal role in facilitating the training of large and complex RL models and await future advancements in this field.

\mypar{Limitations} However, it's important to acknowledge several existing constraints in our proposed system. Firstly, while \sys is specifically designed for large RL models, further research is needed to examine its scalability with increasingly large models and datasets. Secondly, our current focus is on the experience replay segment of the RL pipeline. Consequently, additional investigations are required to understand how our system could be seamlessly integrated with other pipeline segments such as model training and evaluation.

\section*{Acknowledgements}

The authors from Shanghai Jiao Tong University are supported by the National Key R\&D Program of China (2022ZD0114804), Shanghai Municipal Science and Technology Major Project (2021SHZDZX0102), Shanghai Sailing Program (21YF1421900), and National Natural Science Foundation of China (No. 62106141 \& 62076161).



\bibliography{reference}
\bibliographystyle{icml2023}

\newpage
\appendix
\onecolumn


\end{document}